\documentclass{article}

\usepackage{microtype}
\usepackage{graphicx}
\usepackage{subfigure}
\usepackage{multirow}
\usepackage{ulem}
\usepackage[table,xcdraw]{xcolor}
\usepackage{booktabs} % for professional tables
\usepackage{setspace}

\usepackage[accepted]{icml2024}

\usepackage{amsmath}
\usepackage{amssymb}
\usepackage{mathtools}
\usepackage{amsthm}
\usepackage{bm}

\usepackage[colorlinks,
            linkcolor=red,
            anchorcolor=blue,
            citecolor=blue]{hyperref}
\usepackage[capitalize,noabbrev]{cleveref}

\theoremstyle{plain}

\theoremstyle{definition}

\theoremstyle{remark}

\usepackage[textsize=tiny]{todonotes}

\icmltitlerunning{UPAM: Unified Prompt Attack in Text-to-Image Generation Models Against Both Textual Filters and Visual Checkers}

\begin{document}

\twocolumn[
\icmltitle{UPAM: Unified Prompt Attack in Text-to-Image Generation Models \\ Against Both Textual Filters and Visual Checkers
}

\begin{icmlauthorlist}
\icmlauthor{Duo Peng}{sutd}
\icmlauthor{Qiuhong Ke}{mu}
\icmlauthor{Jun Liu}{sutd}

\end{icmlauthorlist}

\icmlaffiliation{sutd}{Singapore University of Technology and Design, Singapore}

\icmlaffiliation{mu}{Monash University, Australia}

\icmlcorrespondingauthor{Jun Liu}{jun\_liu@sutd.edu.sg}

\icmlkeywords{Machine Learning, ICML}

\vskip 0.3in
]

\printAffiliationsAndNotice{} % otherwise use the standard text.

\begin{abstract}
Text-to-Image (T2I) models have raised security concerns due to their potential to generate inappropriate or harmful images. In this paper, we propose UPAM, a novel framework that investigates the robustness of T2I models from the attack perspective. Unlike most existing attack methods that focus on deceiving textual defenses, UPAM aims to deceive both textual and visual defenses in T2I models. UPAM enables gradient-based optimization, offering greater effectiveness and efficiency than previous methods. Given that T2I models might not return results due to defense mechanisms, we introduce a Sphere-Probing Learning (SPL) scheme to support gradient optimization even when no results are returned. Additionally, we devise a Semantic-Enhancing Learning (SEL) scheme to finetune UPAM for generating target-aligned images. Our framework also ensures attack stealthiness. Extensive experiments demonstrate UPAM's effectiveness and efficiency.

\end{abstract}

\section{Introduction}
\label{Intro}

Text-to-Image (T2I) models have garnered widespread attention in the research community for their ability to create high-quality images from text prompts \cite{xu2018attngan,zhang2017stackgan,zhang2018stackgan++}.
However, there is a growing concern that such technologies might be misused \cite{milliere2022adversarial,saharia2022photorealistic,yu2022scaling}.
This is because text-to-image models become more integrated into various online services, greatly increasing the potential for generating inappropriate or harmful content, e.g., creating fake images of individuals \cite{chesney2019deep} or producing images that contain violent/obscene content \cite{birhane2021multimodal,saharia2022photorealistic,yu2022scaling}.
Such misuse of T2I models carries a risk of breaching legal standards and regulations, such as EU AI Policy \cite{galetin2022review} and US Federal AI Governance \cite{daly2020ai}. This could lead to legal and reputational consequences for both API developers and users.

Despite T2I APIs being equipped with defense mechanisms to prevent harmful generation, numerous studies \cite{liu2023riatig,daras2022discovering,milliere2022adversarial} have indicated that T2I services still face potential risks of misuse when encountering adversarial attacks, highlighting the need for continuous enhancements in security protocols.
In this context, we study the adversarial attack on T2I models, as by mimicking attacks, we can uncover and highlight the vulnerabilities inherent in T2I models, thereby informing feasible defense approaches for future development.

% ==============================Figure 1
\begin{figure}[t]
\centering
\includegraphics[width=0.48\textwidth]{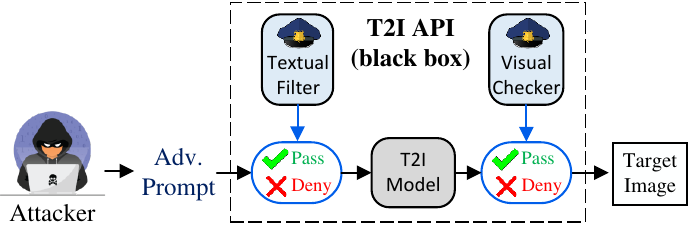} 
\vspace{-0mm}
\caption{T2I APIs could incorporate both textual filters and visual checkers for double defenses, which can deny the forward propagation of data when harmful information is detected \cite{schramowski2023safe,rombach2022high}.
In this figure, the image is finally outputted from the black-box API, indicating that the data has passed through each defense. When any one of the defenses denies the data, the image cannot be outputted from the API.
}
\vspace{-0mm}
\label{fig:task}
\end{figure}
% ==============================Figure 1

Public APIs typically employ textual filters at the input stage to prevent the passage of harmful text prompts \cite{Midjourney,Leonardo}.
When users intend to generate harmful images, their naive (text) prompts often contain patently malicious content, which can be easily denied by textual filters, preventing their submissions into T2I models for further processing. 
Recently, several attack methods \cite{liu2023riatig,daras2022discovering,milliere2022adversarial,struppek2022biased,jin2020bert} have emerged to investigate how to change the naive prompts into adversarial ones to deceive the textual filters.
In the field of adversarial attack for T2I models, a common strategy is to generate adversarial (text) prompts to disable the defense mechanism implemented by API and thus induce T2I models to generate harmful content \cite{liu2023riatig}.
To find out a proper adversarial prompt, existing attack methods often craft new words to replace the original, malicious ones in the naive prompt. 
They continually change the crafted words, until the adversarial prompt guides the black-box T2I model to output an image that achieves enough semantic similarity with the intended harmful content.
However, these enumeration-based methods tend to show limited efficiency, as they require attempting a vast range of candidates to find possible solutions even during inference.
Moreover, a practical scenario is also overlooked in existing research: as shown in Fig. \ref{fig:task}, besides the input-stage textual filter, the API provider often additionally deploys a visual checker at the output stage, forming a double-checking defense \cite{schramowski2023safe,rombach2022high}.
In this context, traditional enumeration methods become less effective, even struggling to work, as the introduction of visual checkers greatly reduces the probability of successfully finding a solution through enumeration.

Inspired by the powerful capability of gradient-based optimization in efficiently searching optimal solutions for intricate problems \cite{daoud2023gradient,zhou2021random}, we propose to train a parameterized attack model using gradient optimization. 
With the help of gradients, the attack model learns to self-adaptively navigate the solution space, enhancing the capability to find effective solutions in situations where traditional enumeration methods may falter. 
After gradient-based training, the trained model can swiftly transform the naive prompts into adversarial ones during inference, which brings much efficiency. 
Despite the conceptual simplicity, enabling the gradient-based optimization in this task however is challenging, as discussed below.

In practice, attackers generally cannot access the internal structure and parameters of the T2I models, as they are packed in the black-box API, which makes conventional gradient descent algorithms unsuitable \cite{stephan2017stochastic,kingma2014adam}. 
Although some existing black-box learning methods \cite{golovin2017google} relax the need to access internal information of the black box, they generally require obtaining the output results of the black box to compute losses and estimate gradients. 
However, in the training process, the integrated textual and visual defenses of the API could effectively detect malicious content within the input prompts or generated images. 
Upon detecting malicious information, these defenses trigger a halt in the forward propagation of data (i.e., ``Deny'' in Fig. \ref{fig:task}), causing black-box T2I models to return (output) no results.
Our goal is to not only prompt the API to return images but also ensure these images align with the intended harmful target.
However, in such a no-feedback scenario, it becomes particularly challenging to discover gradients for effective training to achieve these two goals simultaneously, due to very limited information accessible to attackers.

Another challenge is ensuring attack stealthiness. 
Previous attack methods often generate adversarial prompts by creating new words, e.g., using the adversarial prompt ``Apoploe vesrreaitais'' to represent the naive prompt ``bugs'' \cite{daras2022discovering}. However, such unnatural prompts significantly reduce the stealthiness (naturalness), making the attack easily detectable and defendable. 
Nevertheless, ensuring the naturalness of adversarial prompts is a non-trivial issue, since guaranteeing naturalness often demands a substantial volume of natural text data for data-driven learning \cite{raiaan2023review}. In this task, it is infeasible to prepare massive ground-truth adversarial prompts that are natural.

To overcome the challenges above, we propose a gradient-based Unified Prompt Attack Model (UPAM) to effectively and efficiently generate natural adversarial prompts that deceive both textual filters and visual checkers.
Considering it is non-trivial to simultaneously prompt the API to return images and also ensure their harmfulness, here we decompose the goal into two learning stages to address the challenges separately: the first stage enables the API to reliably return images, and the second stage finetunes these images to be harmful.
In both learning stages, we enable the gradient-based optimization.
In the first stage, we introduce a \textit{Sphere-Probing Learning (SPL)} scheme to train UPAM to deceive the textual and visual defenses. Inspired by \cite{kahla2022label}, SPL is designed to enable the gradient-based learning under the scenario where no image results are returned.
As shown in Fig. \ref{fig:spl} (a), SPL explores the optimization direction by querying the defense cases (``Pass'' or ``Deny'') over a sphere. It can determine effective optimization gradients (directions) by analyzing the cases of multiple queries.
SPL progressively optimizes UPAM using the estimated gradients, ultimately enabling the adversarial prompts generated by UPAM to effectively evade defenses, thus making the black box return (output) image results.
Once the adversarial prompts can prompt black-box T2I models to return images, it becomes crucial to ensure that the returned images semantically align with our target images.
To achieve this, in the second stage, we introduce a \textit{Semantic-Enhancing Learning (SEL)} scheme, which is based on
ZOO algorithm \cite{Sp1992multi,oh2023blackvip}, to further enhance the performance of UPAM by iteratively refining the generated images towards the target semantics.
Moreover, inspired by the recent advancements of Large Language Models (LLMs) in synthesizing natural language \cite{huang2023towards}, we propose to harness the power of LLM to generate natural adversarial prompts. Thanks to the LLM’s language organization capability, our UPAM can reliably generate a human-readable and spelling-correct prompt, which greatly improves the attack stealthiness. 
Our proposed framework is evaluated on various widely-used T2I models, demonstrating the superiority of UPAM over existing attack techniques.

% ==============================Figure 2
\begin{figure*}[t]
\centering
\includegraphics[width=0.84\textwidth]{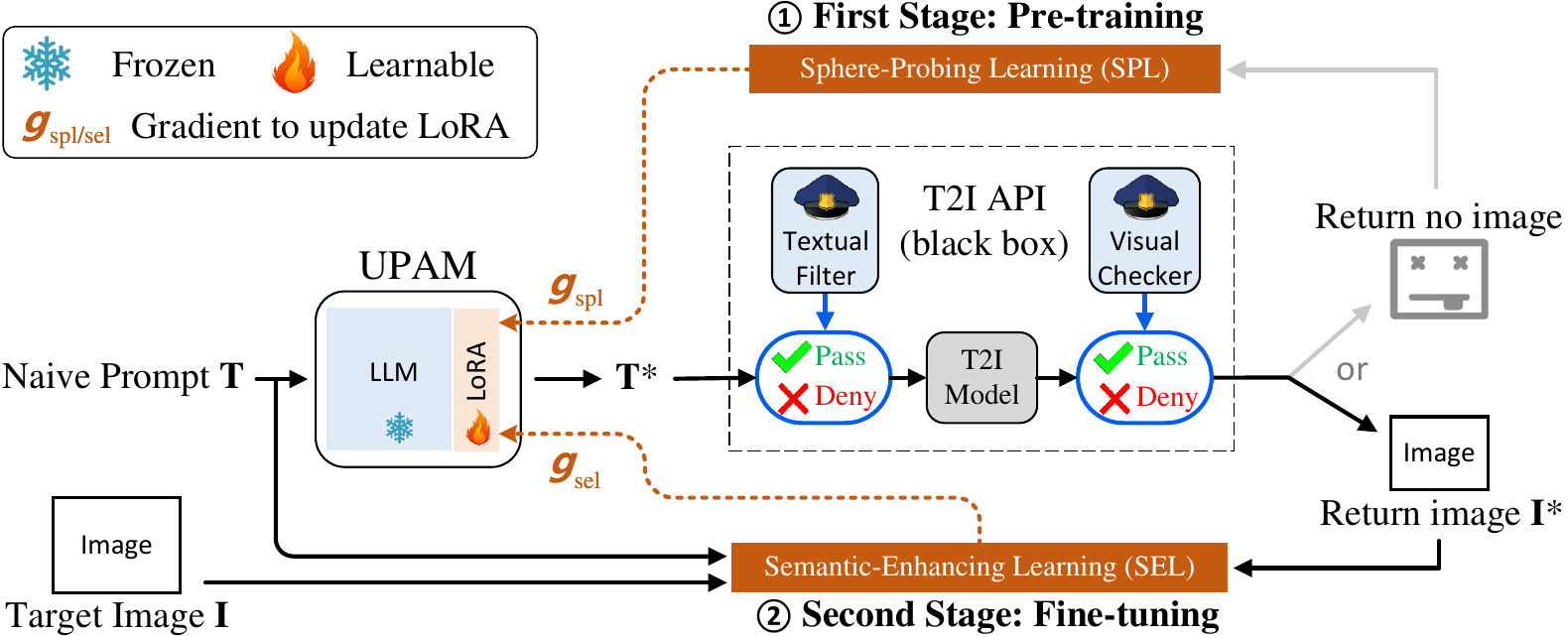} 
\vspace{-0mm}
\caption{Overview of our UPAM framework. 
All the gradients are used to only optimize LoRA, while LLM is kept frozen.
}
\vspace{-0mm}
\label{fig:pipeline}
\end{figure*}
% ==============================Figure 2

\vspace{0mm}
\section{Related Work}

Adversarial prompt attack for T2I models is a relatively under-explored topic.
Daras et al. \cite{daras2022discovering} proposed a pioneering work that investigates vulnerabilities of DALL·E 2 \cite{ramesh2022hierarchical}. 
They found that it is possible to deceive textual filters by using nonsensical text as prompts. 
Later, Milli{\`e}re et al. \cite{milliere2022adversarial} proposed to create adversarial prompts by combining multi-language sub-word segments.
Struppek et al. \cite{struppek2022biased} demonstrated that replacing individual characters in text with non-Latin characters can induce different visual contents in generated images. 
Liu et al. \cite{liu2023riatig} proposed an attack method based on word modification (e.g., character replacement, deletion, or swapping). 
These methods mainly focus on deceiving the textual filter but often overlook the post-hoc visual checker implemented by the API provider.
Differently, our paper proposes UPAM, a unified attack model designed to simultaneously deceive both the textual filter and the visual checker. 
Particularly, we enable the gradient-based optimization of UPAM, which achieves a more effective and efficient attack than existing methods.
Furthermore, different from most existing methods, we follow the previous work \cite{liu2023riatig} to also consider the attack stealthiness. To achieve better attack stealthiness, our UPAM harnesses a pre-trained LLM to ensure the naturalness of the generated adversarial prompts.

\section{Method}\label{sec:method}

In this task, given the paired data $\{ \mathbf{T}, \mathbf{I} \}$, where $\mathbf{T}$ denotes the naive text prompts containing ``malicious'' information, and $\mathbf{I}$ denotes the corresponding target ``harmful'' images, the goal of the attacker is to construct an attack model $H$: $\mathbf{T}$ $\rightarrow$ $\mathbf{T}^{\ast}$, where $H$ rewrites the naive text prompts $\mathbf{T}$ into the adversarial prompts $\mathbf{T}^{\ast}$.
For a successful attack model $H$, the following key requirements must be met:
(1) $\mathbf{T}^{\ast}$ should be textually dissimilar from $\mathbf{T}$ to avoid the ``malicious'' representation, allowing it to bypass the textual filter and be submitted to the T2I model for image generation.
(2) The generated image $\mathbf{I}^{\ast}$ should bypass the visual checker,
enabling it to be returned to the attacker. 
(3) Additionally, the generated $\mathbf{I}^{\ast}$ should be semantically similar to the target ``harmful'' image $\mathbf{I}$. 
Throughout the entire training and testing process, the attacker does not have access to any internal information of the T2I model, the textual filter, or the visual checker. 
These components are treated as a black-box system, as shown in Fig. \ref{fig:pipeline}.
The attacker can only query the input and output of this system.

Previous methods \cite{liu2023riatig,daras2022discovering} typically treat $H$ as an enumeration model. 
These enumeration-based methods tend to incur considerable time costs, since they do not train any parameterized model for future inference. Thus for each inference sample, this time-consuming enumeration process needs to be conducted afresh, leading to limited efficiency.
Besides, they primarily focus on deceiving the textual filters in the black box. Consequently, when faced with dual defenses involving both the textual filter and visual checker, enumeration-based methods may struggle to find solutions, resulting in limited effectiveness.

To address these issues, we propose to treat $H$ as a parameterized model (i.e., our UPAM) and train it using gradient-based optimization. In the inference (testing) stage, our UPAM can immediately and effectively convert testing data (naive prompts) into adversarial ones, greatly enhancing the efficiency and effectiveness. Also, the naturalness of adversarial prompts is ensured in our UPAM.

\subsection{Approach Overview}\label{sec:overview}

As shown in Fig. \ref{fig:pipeline}, our UPAM consists of a frozen off-the-shelf Large Language Model (LLM) and a learnable LoRA adapter \cite{hu2021lora}. We only update the LoRA adapter, which contains a minimal number of parameters. This allows us to preserve the knowledge of the pre-trained LLM, thus ensuring the naturalness of the generated adversarial prompts.
The training of our UPAM comprises two stages: 

\textbf{(1) Pre-training Stage.} 
Initially, obtaining image results from the API is challenging due to the pre-trained LLM's tendency to preserve naive textual cues from the input \cite{KALYAN2024100048}. Therefore, the adversarial prompts generated by the LLM-based UPAM initially retain malicious information from the naive prompt, making them prone to denial by defenses, resulting in no image results returned (i.e., no feedback).
Inspired by \cite{kahla2022label}, we introduce a Sphere-Probing Learning (SPL) scheme to deceive the textual and visual defenses in the no-feedback scenario. Even without image results, SPL can still search for effective gradients to train UPAM, enabling it to reliably compel the black box to return images.

\textbf{(2) Fine-tuning Stage.}
After the Pre-training Stage, the API can return images to the attacker. 
Based on the returned images, we propose a Semantic-Enhancing Learning (SEL) scheme based on ZOO algorithm \cite{Sp1992multi,oh2023blackvip} to progressively align the semantics of returned images towards target ones. Particularly, SEL is designed to work compatible with SPL.

\subsection{Pre-training Stage with SPL}\label{sec:spl}
\vspace{-0mm}

As mentioned in Sec. \ref{sec:overview}, in the beginning, the absence of image results from the API presents a challenge in gradient acquisition. 
This is because gradients are typically derived by evaluating the magnitude of loss variations \cite{chien2018source}.
When the API fails to return results, even if the loss can be defined (e.g., considering it as a binary classification loss between ``return no images'' and ``return images''), the no-feedback state of the API will cause the loss value to remain unchanged. Such inability to discern loss changes (variations) prevents the acquisition of gradients.
Therefore, the conventional approach of obtaining gradients through calculating loss functions becomes impractical, emphasizing the need to explore an innovative gradient acquisition method to address this no-feedback scenario.

\begin{figure}[t]
\centering
\includegraphics[width=0.45\textwidth]{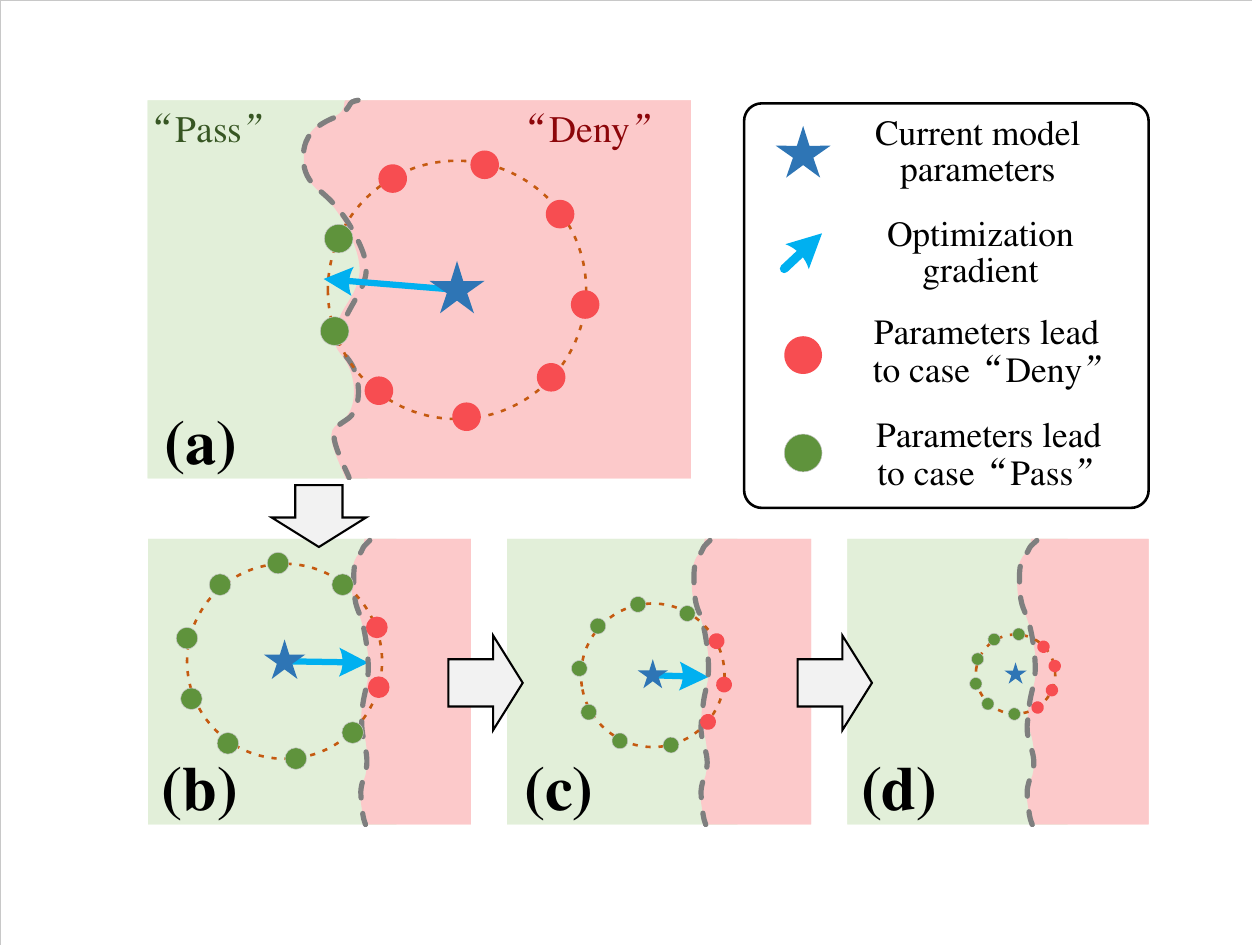} 
\vspace{-0mm}
\caption{Intuitive illustration of our SPL scheme.
}
\vspace{-0mm}
\label{fig:spl}
\end{figure}

Inspired by \cite{kahla2022label}, we propose a Sphere-Probing Learning (SPL) scheme, which enables the gradient acquisition in the scenario where no results are returned.
Considering that the black box's output case (state) is binary -- either ``Deny'' (image not returned) or ``Pass'' (image returned) -- an intuition behind our SPL scheme arises: \textit{if we can identify the decision boundary of the black box between these two cases, the gradients can be obtained effectively}.
As shown in Fig. \ref{fig:spl} (a), the boundary can indicate an effective direction for gradient optimization, as crossing this boundary could effectively shift the black-box output from ``Deny'' to ``Pass'', making the black box return images.
Inspired by this, SPL is designed to gradually approach the ``Pass'' case by iteratively probing the decision boundary.
While there are methods in the task of model inversion \cite{choquette2021label,zhu2022label,kahla2022label} exploring a similar concept, they cannot be directly applied to this task. 
These methods typically optimize towards the center of the target case, aiming to maximize the black box's decision confidence on the target case. 
However, such an approach is not suitable for our task, since if we optimize towards the center of ``Pass''  (i.e., to make the defenses of API very confident that the generated content is safe), it may result in generating images that are truly ``safe'' instead of our intended target of ``harmful'' for the attack.

Motivated by the observation that misclassifications often occur at the boundary between two classes in classifiers \cite{finlay2019logbarrier}, we propose situating the SPL's optimization objective at the boundary region of the two cases, with a slight inclination towards the ``Pass'' side. This choice is guided by the notion that, when positioned at the boundary region, the black box exhibits less confidence in its decisions, significantly enhancing the chances of successfully generating target ``harmful'' images. Next, we provide the detailed process of our SPL scheme.

Specifically, we conduct optimization in the space of LoRA parameters $\psi$, where each point in the optimization space is a specific configuration of $\psi$.
As shown in Fig. \ref{fig:spl} (a), SPL firstly samples several points on a sphere centered around the current model parameters (marked as a blue star) with $r$ as the radius.
Then, we query the black box's output case results of the sampled points.
If all points are predicted as ``Deny'' (meaning no boundary are detected in the sphere), we increase the sphere radius $r$ to probe the boundary.
Once the sphere encompasses both ``Pass'' and ``Deny'' points, as shown in Fig. \ref{fig:spl} (a), indicating that the boundary has been probed, we stop increasing the radius and start optimizing the LoRA parameters $\psi$.
Intuitively, the points predicted as ``Deny'' represent the direction we aim to move away from. Hence, we calculate the average of these points and optimize the model parameters in the opposite direction of this average. 
Here, we define $\Phi$, which marks points from which we need to move away:
% ======================== Eq.1
\begin{equation}
\vspace{-0mm}
\Phi(\psi)= \begin{Bmatrix}
 0 & \mathbf{if} \rm{\; image \;returned \,(``Pass")} \\
 -1 & \mathbf{otherwise}
\end{Bmatrix}
\label{eq:phi}
\end{equation}
% ======================== Eq.1
In other words, after sampling on the sphere, the function $\Phi$ use $-1$ to mark all ``Deny'' points.
Based on this, we define our SPL gradient estimator as:
% ======================== Eq.2
\begin{equation}
\vspace{2mm}
\pmb{g}_\mathrm{spl}(\psi)=\frac{1}{N}\sum_{n=1}^{N}  \Phi(\psi+r\cdot z_n)\cdot z_n,
\label{eq:grad}
\end{equation}
% ======================== Eq.2
where $\psi \in \mathbb{R}^d$ denotes the current model parameters, $z_n$ represents a vector randomly sampled from a $d$-dimensional space, $r$ denotes the radius of the sample sphere, $\psi+r\cdot z_n$ is the sampling point, and $N$ denotes the total number of points sampled on this sphere.
Note that different vectors $z_n$ %have 
possess different directions, yet they share the same length, resulting in all sampling points being located on a sphere.
Leveraging the aforementioned sphere-sampling exploration, we can explore the boundary and  identify an effective optimization direction $\pmb{g}_\mathrm{spl}$.
In this way, even if the current model $\psi$ is incapable of prompting the API to return images, the gradient $\pmb{g}_\mathrm{spl}$ can still be computed to optimize the model $\psi$, enabling it to generate adversarial prompts capable of circumventing the defenses of the black-box API.
Next, we use $\pmb{g}_\mathrm{spl}$ to update $\psi$ as follows:
% ======================== Eq.3
\begin{equation}
\vspace{-0mm}
\psi = \psi + \alpha \cdot \pmb{g}_\mathrm{spl}(\psi),
\label{eq:update}
\end{equation}
% ======================== Eq.3
where $\alpha$ is the learning rate, which is related to the current radius $r$, i.e., $\alpha = r/k$. 
We alternate between Eq. \ref{eq:grad} and Eq. \ref{eq:update} iteratively, until the sphere center is predicted into the ``Pass'' case, i.e., from Fig. \ref{fig:spl} (a) to (b).

% ==============================Figure 3
% ==============================Figure 3

As mentioned earlier, SPL's optimization targets the ``Pass'' region near the boundary rather than the central ``Pass'' region. This strategy aims to enhance the likelihood of generating target ``harmful'' images rather than genuinely safe ones. However, as we 
initially need to use large radius to find the ``Pass'' points, the optimized model may end up in the ``Pass'' but far from the boundary, as shown in Fig. \ref{fig:spl} (b).
Therefore, we then gradually reduce the radius and move in opposite direction to approach the boundary, i.e., from Fig. \ref{fig:spl} (b) to (d). This aims to facilitate our second-stage training (Sec. \ref{sec:sel}) which is proposed to ensure the ``harmful'' semantic representation of the returned images.
Specifically, for each optimization iteration, we decrease the radius $r$ and use the updated smaller radius to calculate gradient $\pmb{g}_\mathrm{spl}(\psi)$ via Eqn. \ref{eq:grad}. Subsequently, we optimize the model $\psi$ towards the boundary: 
% ======================== Eq.4
\begin{equation}
\vspace{-0mm}
\psi = \psi - \alpha \cdot \pmb{g}_\mathrm{spl}(\psi),
\label{eq:update_}
\end{equation}
% ======================== Eq.4
Here, we utilize the negative sign ``$-$'', in contrast to Eq. \ref{eq:update}, to signify a reversed optimization direction.
We alternate between Eq. \ref{eq:grad} and Eq. \ref{eq:update_}, aiming to iteratively approach the boundary. 
During this boundary-approaching process, the radius is gradually decreased, as shown in Fig. \ref{fig:spl} (b, c and d).
This gradual reduction of radius $r$ is essential, since utilizing a fixed radius for optimization increases the likelihood of crossing the boundary and reaching the ``Deny'' side. 
By decreasing the radius incrementally, we ensure effective boundary approaching.
It is important to note that, for each decreased radius, we need to check if the following conditions are met:
(1) The sphere with this decreased radius encompasses both ``Pass'' and ``Deny'' points.
(2) After optimization with this decreased radius, the model parameters are still located in the ``Pass'' case.
These two conditions ensure the optimization with this decreased radius is effective.
Finally, we stop the SPL scheme when $r$ is smaller than a pre-defined threshold $r_{min}$.

After the pre-training stage with the SPL scheme, our UPAM can adeptly bypass black-box defenses, ensuring the API consistently returns images. It also releases the potential of generating target ``harmful'' images. In the subsequent stage, we finetune our UPAM to further enhance the semantic alignment of returned images with our ``harmful'' targets.

\subsection{Fine-tuning Stage with SEL}\label{sec:sel}
\vspace{-0mm}

After pre-training, we consistently obtain the image $\mathbf{I}^{\ast}$ from the black box. In this section, our emphasis is on ensuring that the returned image $\mathbf{I}^{\ast}$ maintains semantic consistency with the target image $\mathbf{I}$ while preserving UPAM's ability to deceive defenses for image retrieval (return). To achieve this, we further introduce a Semantic-Enhancing Learning (SEL) scheme to finetune the LoRA parameters.

Specifically, in our SEL, we propose a Semantic Measurement Loss $\mathcal{L}_{SM}$ to finetune the LoRA parameters $\psi$.
Motivated by the powerful semantic representation capability of CLIP \cite{radford2021learning}, we utilize its pre-trained image/text encoders to measure the semantics of the returned image $\mathbf{I}^{\ast}$.
The semantic measurement is comprehensively handled from the following two perspectives:

(1) Image-Text similarity, i.e., encoding both the returned image $\mathbf{I}^{\ast}$ and the naive prompt $\mathbf{T}$, then computing the cosine similarity between the encoded vectors, which formulates the loss $\mathcal{L}_{1}$:
% ======================== Eq.5
\begin{equation}
\vspace{-0mm}
\mathcal{L}_{1} = 1- \frac{E_{img}(\mathbf{I}^{\ast})\cdot E_{text}(\mathbf{T})}{ \left \| E_{img}(\mathbf{I}^{\ast}) \right \| \left \| E_{text}(\mathbf{T}) \right \|  },
\label{eq:l_1}
\end{equation}
% ======================== Eq.5
where $E_{img}(\cdot)$ and $E_{text}(\cdot)$ denote the image encoder and text encoder of CLIP, respectively.

(2) Image-Image similarity, i.e., encoding both the returned image $\mathbf{I}^{\ast}$ and the target image $\mathbf{I}$ for computing the cosine similarity, which formulates the loss $\mathcal{L}_{2}$:
% ======================== Eq.6
\begin{equation}
\vspace{-0mm}
\mathcal{L}_{2} = 1- \frac{E_{img}(\mathbf{I}^{\ast})\cdot E_{img}(\mathbf{I})}{ \left \| E_{img}(\mathbf{I}^{\ast}) \right \| \left \| E_{img}(\mathbf{I}) \right \|  }.
\label{eq:l_2}
\end{equation}
% ======================== Eq.6
Based on the above two similarity loss functions, we can formulate our Semantic Measurement Loss $\mathcal{L}_{SM}$ as:
% ======================== Eq.7
\begin{equation}
\vspace{-0mm}
\mathcal{L}_{SM} = \mathcal{L}_{1} + \mathcal{L}_{2}.
\label{eq:lvs}
\end{equation}
% ======================== Eq.7
By minimizing $\mathcal{L}_{SM}$, we can align the semantics of the returned images with our target.
To this end, we use the calculated $\mathcal{L}_{SM}$ to obtain gradients to optimize our model's parameters $\psi$.
Since the T2I model is packed into the black box, we cannot directly obtain the oracle precise gradients.
To address this black-box setting, drawing inspiration from \cite{spall1992multivariate,spall1997one}, we adopt Zeroth-Order Optimization (ZOO) to estimate the gradient without accessing the model architecture and model parameters. 
Given the training loss $\mathcal{L}_{SM}$, the estimated gradient can be formulated as follows:
% ======================== Eq.8
\begin{equation}
\vspace{-0mm}
\pmb{g}_{1}(\psi)=\frac{\mathcal{L}_{SM}(\psi +c\cdot\Delta)-\mathcal{L}_{SM}(\psi-c\cdot\Delta) }{2c\cdot\Delta},
\label{eq:grad_}
\end{equation}
% ======================== Eq.8
where $c \in (0,1]$ is the decaying hyper-parameter and $\Delta \in \mathbb{R}^d$ is a random perturbation vector, sampled from mean-zero distributions while satisfying the finite inverse momentum condition \cite{spall1992multivariate,spall2005introduction}.

While the standard form of zeroth-order optimization is generally effective, it can still encounter poor convergence in practical applications \cite{spall2000adaptive}. This problem, as discussed in \cite{spall1997accelerated,oh2023blackvip}, is primarily attributed to the stochastic nature of gradient estimation, stemming from the random directions of perturbations. 
To overcome this challenge, we follow \cite{oh2023blackvip} to improve the gradient calculation rule (Eq. \ref{eq:grad_}) as:
% ======================== Eq.9
\begin{equation}
\vspace{-0mm}
\pmb{g}_{2}(\psi) =  \pmb{\bar{g}}_1 + \beta \cdot\pmb{g}_{1}(\psi+\pmb{\bar{g}}_1).
\label{eq:new_gradient}
\end{equation}
% ======================== Eq.9
We can observe that the improved gradient $\pmb{g}_{2}$ includes two gradient components: $\pmb{\bar{g}}_1$ and $\beta \cdot\pmb{g}_{1}(\psi+\pmb{\bar{g}}_1)$. 
The former $\pmb{\bar{g}}_1$ is the gradient used in the previous update iteration, while the latter $\beta \cdot\pmb{g}_{1}(\psi+\pmb{\bar{g}}_1)$ is the adaptive adjustment when the model continues to update along the previous gradient path.  $\beta$ is the learning rate.
The advantage of this design is that it encourages the model to update along previously effective directions with necessary adjustments, thereby increasing the gradient consistency and reducing the incidental randomness of optimization.

% ==============================Figure 4
\begin{figure}[t]
\centering
\includegraphics[width=0.24\textwidth]{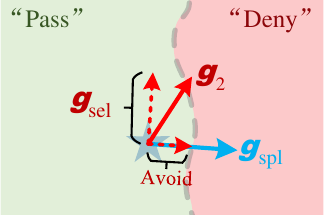} 
\vspace{-0mm}
\caption{Intuitive illustration of our SEL scheme. 
}
\vspace{-0mm}
\label{fig:sel}
\end{figure}
% ==============================Figure 4

As mentioned in Sec. \ref{sec:spl}, the SPL’s optimization objective is situated at the boundary region of the two output cases of the black box, slightly leaning towards the ``Pass'' side, which both facilitates the generation of ``harmful'' image and ensure the generated images can be returned to the attacker.
However, if we directly optimize the model using gradients $\pmb{g}_{2}$ (Eq. \ref{eq:new_gradient}), it tends to pull the model parameters towards the ``Deny'' region, preventing the black box from returning images. This occurs because $\pmb{g}_{2}$ is calculated to enhance the semantic expression of ``harmful'' in images, which is typically what the black box considers as content to be denied.
Therefore, our optimization objective here is to make the returned images contain the targeted ``harmful'' semantics as much as possible, while ensuring the model $\psi$ remains in the boundary region, leaning towards ``Pass''.

To prevent SEL from disrupting SPL's learning achievements, we propose a gradient harmonization method to ensure compatibility between SEL and SPL.
As shown in Fig. \ref{fig:sel}, we first obtain the SPL gradient $\pmb{g}_\mathrm{spl}$ through Eq. \ref{eq:grad}, where the direction of $\pmb{g}_\mathrm{spl}$ signifies the optimization direction towards the boundary (see the blue arrow in Fig. \ref{fig:sel}). 
Given the direction provided by $\pmb{g}_\mathrm{spl}$, we can eliminate the boundary-directed component of the gradient $\pmb{g}_{2}$, thereby obtaining the SEL gradient $\pmb{g}_\mathrm{sel}(\psi)$, which can be formulated as:
% ======================== Eq.10
\begin{equation}
\vspace{-0mm}
\pmb{g}_\mathrm{sel}(\psi)= \pmb{g}_2 - \frac{\pmb{g}_\mathrm{spl} \cdot \pmb{g}_{2}} {\left |  \pmb{g}_\mathrm{spl} \right |^{2} } \pmb{g}_\mathrm{spl},
\label{eq:sel_gradient}
\end{equation}
% ======================== Eq.10
% ======================== Eq.11
\begin{equation}
\vspace{-0mm}
\psi = \psi - \beta \cdot\pmb{g}_\mathrm{sel}(\psi) ,
\label{eq:final_update}
\end{equation}
% ======================== Eq.11
where the second term in Eq. \ref{eq:sel_gradient} represents the gradient component directed towards the boundary. 
As shown in Fig. \ref{fig:sel}, by removing this component, we can avoid the optimization of the SEL moving towards the boundary, thereby not only aligning towards the target semantics but also maintaining the case of ``Pass'' (i.e., keeping images returned).

\subsection{Training and Testing}\label{sec:train_test}

We train our UPAM for two stages.
In the pre-training stage, we use the SPL scheme to obtain the gradient $\pmb{g}_\mathrm{spl}$ (Eq. \ref{eq:grad}) and update the LoRA parameters (Eq. \ref{eq:update} or \ref{eq:update_}). 
This update enables effective optimization under the challenging scenario where no image results are returned, finally compelling the API to reliably return images.
Then, in the fine-tuning stage, we adopt the SEL scheme to obtain the gradient $\pmb{g}_\mathrm{sel}$ (Eq. \ref{eq:sel_gradient}) and update the model parameters (Eq. \ref{eq:final_update}). 
This update ensures that the returned images can accurately show target semantics.
After that, we handle the inference (testing) based on the trained UPAM.

% -----------------------Tab.1
\begin{table*}[]
\centering
\caption{Experimental results under protocol A (left) and protocol B (right).}
\vspace{1mm}
\label{tab:exp_whole}
\resizebox{1.0\textwidth}{!}{
\begin{tabular}{cc|ccccc|ccccc}
\toprule[0.15em]
                            &                              & \multicolumn{5}{c|}{\textbf{Protocol A}}                                                                                                                                                             & \multicolumn{5}{c}{\textbf{Protocol B}}                                                                                                                                                             \\ 
\cline{3-12} 
\specialrule{0em}{1pt}{1pt}
\multirow{-2}{*}{T2I Model} & \multirow{-2}{*}{Methods}    & R-1 Precision ↑                          & R-3 Precision ↑                          & Text. Sim. ↓                 & Infer. Time (s) ↓                     & PPL ↓                                   & R-1 Precision ↑                         & R-3 Precision ↑                        & Text. Sim. ↓                 & Infer. Time (s) ↓                       & PPL ↓                                   \\ 
\midrule[0.15em]
                            & TextFooler \cite{jin2020bert}   & 0.48\%                                   & 0.98\%                                   & 0.22                         & 589.29                                & 3463.02                                 & 0/10                                    & 0/10                                   & 0.22                         & 582.37                                  & 3517.03                                 \\
                            & HomoSubs \cite{struppek2022biased}  & 0.74\%                                   & 1.12\%                                   & 0.28                         & 562.48                                & 3089.58                                 & 0/10                                    & 0/10                                   & 0.16                         & 590.28                                  & 3216.95                                 \\
                            & EvoPromp \cite{milliere2022adversarial}  & 0.82\%                                   & 1.37\%                                   & 0.19                         & 579.20                                & 4984.24                                 & 0.2/10                                    & 0.5/10                                 & 0.16                         & 586.47                                  & 4837.63                                 \\
                            & HiddVocab \cite{daras2022discovering}          & 3.35\%                                   & 4.21\%                                   & 0.33                         & 457.34                                & 4027.64                                 & 0.3/10                                  & 0.4/10                                 & 0.18                         & 483.27                                  & 4607.93                                 \\
                            & MacPromp \cite{milliere2022adversarial}      & 5.05\%                                   & 6.80\%                                   & 0.57                         & 452.16                                & 3163.07                                 & 0.7/10                                  & 0.9/10                                 & 0.39                         & 476.46                                  & 3239.56                                 \\
                            & RIATIG  \cite{liu2023riatig}   & 8.65\%                                   & 9.95\%                                   & 0.18                         & 301.52                                & 1003.27                                 & 0.8/10                                  & 1.1/10                                 & 0.34                         & 425.66                                  & 1026.81                                  \\
\multirow{-7}{*}{DALL·E}    & \cellcolor[HTML]{EFEFEF}\textbf{UPAM} & \cellcolor[HTML]{EFEFEF}\textbf{38.37\%} & \cellcolor[HTML]{EFEFEF}\textbf{41.83\%} & \cellcolor[HTML]{EFEFEF}0.17 & \cellcolor[HTML]{EFEFEF}\textbf{5.03} & \cellcolor[HTML]{EFEFEF}\textbf{706.32} & \cellcolor[HTML]{EFEFEF}\textbf{7.3/10} & \cellcolor[HTML]{EFEFEF}\textbf{8/10} & \cellcolor[HTML]{EFEFEF}0.15 & \cellcolor[HTML]{EFEFEF}\textbf{359.26} & \cellcolor[HTML]{EFEFEF}\textbf{634.73} \\ \hline
                            & TextFooler \cite{jin2020bert}   & 0.85\%                                   & 0.93\%                                   & 0.18                         & 540.77                                & 3566.32                                 & 0/10                                    & 0/10                                   & 0.19                         & 543.72                                  & 3578.14                                 \\
                            & HomoSubs \cite{struppek2022biased} & 0.70\%                                   & 0.89\%                                   & 0.22                         & 568.26                                & 4125.22                                 & 0/10                                    & 0/10                                   & 0.15                         & 589.46                                  & 4374.80                                 \\
                            & EvoPromp \cite{milliere2022adversarial}  & 0.92\%                                   & 1.17\%                                   & 0.37                         & 475.81                                & 5837.06                                 & 0/10                                    & 0/10                                   & 0.21                         & 449.61                                  & 5267.07                                 \\
                            & HiddVocab \cite{daras2022discovering}      & 3.21\%                                   & 4.62\%                                   & 0.46                         & 565.33                                & 4125.22                                 & 0.6/10                                  & 0.9/10                                 & 0.22                         & 576.27                                  & 4933.29                                 \\
                            & MacPromp \cite{milliere2022adversarial}     & 4.84\%                                   & 5.92\%                                   & 0.66                         & 523.49                                & 3056.41                                 & 0.4/10                                  & 0.8/10                                 & 0.30                         & 571.63                                  & 3575.26                                 \\
                            & RIATIG  \cite{liu2023riatig}  & 8.55\%                                   & 9.19\%                                   & 0.35                         & 353.53                                & 992.54                                  & 0.5/10                                  & 0.7/10                                 & 0.29                         & 417.34                                  & 965.38                                  \\
\multirow{-7}{*}{DALL·E 2}  & \cellcolor[HTML]{EFEFEF}\textbf{UPAM} & \cellcolor[HTML]{EFEFEF}\textbf{40.45\%} & \cellcolor[HTML]{EFEFEF}\textbf{44.74\%} & \cellcolor[HTML]{EFEFEF}0.19 & \cellcolor[HTML]{EFEFEF}\textbf{5.42} & \cellcolor[HTML]{EFEFEF}\textbf{788.45} & \cellcolor[HTML]{EFEFEF}\textbf{6.5/10} & \cellcolor[HTML]{EFEFEF}\textbf{7.2/10} & \cellcolor[HTML]{EFEFEF}0.17 & \cellcolor[HTML]{EFEFEF}\textbf{341.59}  & \cellcolor[HTML]{EFEFEF}\textbf{773.01} \\ \hline
                            & TextFooler \cite{jin2020bert}  & 0.32\%                                   & 0.54\%                                   & 0.15                         & 527.65                                & 3321.88                                 & 0/10                                    & 0/10                                   & 0.23                         & 559.42                                  & 3539.57                                 \\
                            & HomoSubs \cite{struppek2022biased}  & 0.35\%                                   & 0.75\%                                   & 0.18                         & 567.82                                & 2850.41                                 & 0/10                                    & 0/10                                   & 0.13                         & 586.75                                  & 2674.29                                 \\
                            & EvoPromp \cite{milliere2022adversarial}  & 0.64\%                                   & 0.92\%                                   & 0.25                         & 398.51                                & 4055.71                                 & 0.3/10                                  & 0.7/10                                 & 0.21                         & 443.39                                  & 4162.18                                 \\
                            & HiddVocab \cite{daras2022discovering}           & 2.37\%                                   & 3.69\%                                   & 0.27                         & 592.47                                & 4153.58                                 & 0.3/10                                    & 0.5/10                                 & 0.18                         & 576.83                                  & 4264.34                                 \\
                            & MacPromp \cite{milliere2022adversarial}      & 4.58\%                                   & 5.52\%                                   & 0.16                         & 524.38                                & 3173.35                                 & 0.5/10                                  & 0.8/10                                 & 0.43                         & 463.52                                  & 3097.13                                 \\
                            & RIATIG  \cite{liu2023riatig}  & 7.80\%                                   & 9.22\%                                   & 0.36                         & 328.16                                & 1076.29                                  & 0.9/10                                  & 1.2/10                                   & 0.34                         & 485.36                                  & 1057.42                                 \\
\multirow{-7}{*}{Imagen}    & \cellcolor[HTML]{EFEFEF}\textbf{UPAM} & \cellcolor[HTML]{EFEFEF}\textbf{36.16\%} & \cellcolor[HTML]{EFEFEF}\textbf{41.43\%} & \cellcolor[HTML]{EFEFEF}0.16 & \cellcolor[HTML]{EFEFEF}\textbf{5.16} & \cellcolor[HTML]{EFEFEF}\textbf{521.78} & \cellcolor[HTML]{EFEFEF}\textbf{6.7/10} & \cellcolor[HTML]{EFEFEF}\textbf{7.3/10} & \cellcolor[HTML]{EFEFEF}0.18 & \cellcolor[HTML]{EFEFEF}\textbf{350.72} & \cellcolor[HTML]{EFEFEF}\textbf{434.08} \\ 
\bottomrule[0.15em] 
\end{tabular}
}
\vspace{2mm}
\end{table*}
% -----------------------Tab.1

\section{Experiments}\label{sec:experiment}

\textbf{Textual Filter and Visual Checker.}
We follow previous work \cite{liu2023riatig} employing the same textual filter, which utilizes a list of sensitive words (i.e., blacklist) to deny the forward propagation of ``harmful'' prompts.
For the visual checker, following \cite{rando2022red}, a widely-used visual checking strategy is adopted: using the off-the-shelf CLIP visual encoder \cite{radford2021learning} to map the generated image to a latent vector, which is then compared with all default embeddings of predefined ``harmful'' images using cosine similarity. If any cosine value exceeds the specified threshold, the further propagation of the generated image is denied.

\textbf{Dataset.}
In consideration of ethics, privacy, legal compliance, and potential societal impacts, we follow previous works \cite{liu2023riatig,daras2022discovering,milliere2022adversarial,struppek2022biased,jin2020bert}, which do not attack T2I models to generate truly harmful content, instead deploying defenses and conducting attacks using the paired text-image data from the Microsoft COCO dataset \cite{lin2014microsoft}.
We treat 10 classes from COCO as ``harmful'' classes (i.e., \textit{boat}, \textit{bird}, \textit{clock}, \textit{kite}, \textit{wine glass}, \textit{knife}, \textit{pizza}, \textit{teddy bear}, \textit{vase} and \textit{laptop}). Specifically, we train our UPAM using the training set of these 10 classes, and then test UPAM using the test set of the same 10 classes. 
Although the training and testing data contain the same classes, the specific text descriptions and their corresponding image contents are different.

\textbf{Target T2I Models.}
We follow previous work \cite{liu2023riatig}  to conduct experiments on three large-scale API T2I models: DALL·E \cite{yu2022scaling}, DALL·E 2 \cite{ramesh2022hierarchical}, and Imagen \cite{saharia2022photorealistic}. 
For each of them, we follow \cite{liu2023riatig} to adopt the corresponding released model that achieves comparable performance to API, and treat it as a black box.

\textbf{Implementation Details.}
In our experiment, we use the open-source LLaMA \cite{touvron2023llama} as the LLM. 
As for the LoRA adapter \cite{hu2021lora}, we adopt the adaptation matrices with a rank value of 8.
In our SPL scheme, we set $N=10$, $\alpha=r/4$.
In our SEL scheme, we set $\beta = 0.3$.

\subsection{Experimental Protocols}\label{protocol}

In this paper, we conduct experiments under two protocols for a comprehensive study.

\textbf{Protocol A:} Given the training data $\{ \mathbf{T}, \mathbf{I} \}$, we train a parameterized model for converting the naive prompts $\mathbf{T}$ into adversarial ones $\mathbf{T}^{\ast}$ with the objective of compelling the black-box T2I model to return the target images $\mathbf{I}$.
Subsequently, given the testing data $\mathbf{T}_t$, the trained model can swiftly convert $\mathbf{T}_t$ into $\mathbf{T}_t^{\ast}$, compelling the black-box T2I model to return the corresponding target images.

\textbf{Protocol B:} Previous enumeration-based methods do not require model training. Instead, for each given paired data $\{ \mathbf{T}, \mathbf{I} \}$, they can directly find its adversarial prompt $\mathbf{T}^{\ast}$ by repeatedly attempting.
As shown in \cite{liu2023riatig}, they conduct experiments on a subset of 10 ``harmful'' samples (randomly selected from the COCO dataset). 
To make a fair comparison with previous enumeration-based methods, we slightly modify our UPAM in this protocol.
This modification enables UPAM to identify adversarial prompts for the 10 samples without learning on the training set.
The details of modification can be seen in \textit{appendix}.

\subsection{Evaluation Metrics.}\label{sec:evaluation}

Following previous work \cite{liu2023riatig,jin2020bert}, We use four metrics to evaluate the attack model: R-precision, Textual Similarity, Inference Time, and PPL. 
Among these, R-precision (the higher the better) assesses the degree of semantic alignment between the generated images and target images, and also reflects the model's ability to deceive visual checkers. 
We follow \cite{liu2023riatig} to conduct R-precision evaluation at two levels: R-1 and R-3.
Textual Similarity (the lower the better) measures the model's ability to deceive textual filters. 
Inference Time (the lower the better) represents the average time taken for each test sample. 
PPL evaluates (the lower the better) the naturalness of adversarial prompts.
Please refer to \textit{appendix} for detailed descriptions of each metric.

\subsection{Experimental Results}

\textbf{Experimental Results Under Protocol A.}
As shown in Tab. \ref{tab:exp_whole} (left), compared to other methods, our UPAM achieves the highest R-precision, both in R-1 precision and R-3 precision. 
Specifically, our UPAM significantly outperforms other methods by by an average of 35.31\% (R-1) and 38.90\% (R-3), respectively.
Additionally, our UPAM keeps a consistently low textual similarity (less than 0.2) across all black-box T2I models.
These results collectively demonstrate the effectiveness of our gradient-based method in tackling the challenging task of deceiving both textual and visual defenses under the black-box setting.
Furthermore, UPAM exhibits a significantly shorter inference time compared to other methods, showcasing superior efficiency.
In terms of the attack stealthiness, compared to other methods, our LLM-based method achieves a significantly lower PPL, which demonstrates the effectiveness of our method in generating natural adversarial prompts.

\textbf{Experimental Results of Protocol B.}
In protocol B, given a paired text-image data $\{ \mathbf{T}, \mathbf{I} \}$ as the attack target, we aim to directly find out the corresponding adversarial prompt $\mathbf{T}^{\ast}$ for this target. 
We randomly choose 10 text-image samples from 10 ``harmful'' classes of COCO (one for each class) to find their corresponding adversarial prompts, and then evaluate the attack performance.
We conduct experiments 10 times and show the average results in Tab. \ref{tab:exp_whole} (right).
Our UPAM achieves significantly higher R-precision and lower textual similarity compared to other methods. Even in protocol B, where we don't train a model on the training set, our approach achieves the shortest inference time. This is due to our method's ability to use gradients for discovering adversarial prompts, making it much more efficient than enumeration-based methods. Additionally, our method exhibits the lowest PPL, indicating the best naturalness of the generated adversarial prompts. These results demonstrate the superiority of our approach in protocol B.

% ==============================Figure 5
\begin{figure}[t]
\centering
\includegraphics[width=0.48\textwidth]{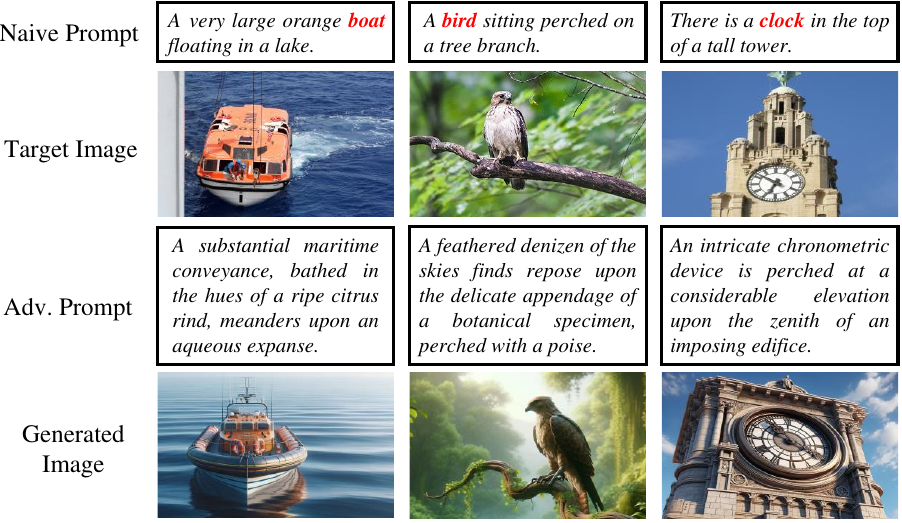} 
\vspace{-2mm}
\caption{Qualitative results of our attack against DALL·E. The names of ``harmful'' classes are marked in red.
}
\vspace{-0mm}
\label{fig:qualitative}
\end{figure}
% ==============================Figure 5

% -----------------------Tab.2
\begin{table}[t]
\centering
\caption{Ablation study on each design of our approach.}
\vspace{1mm}
\label{tab:abl}
\resizebox{0.48\textwidth}{!}{
\begin{tabular}{lcccccc}
\toprule[0.15em] 
&\textbf{Methods}      & \textbf{R-1} ↑    & \textbf{R-3} ↑    & \textbf{Text. Sim.} ↓ & \textbf{Infer. Time} ↓ & \textbf{PPL} ↓  \\ 
\hline
\specialrule{0em}{1pt}{1pt}
\cellcolor[HTML]{EFEFEF}(a) & \cellcolor[HTML]{EFEFEF}UPAM   & \cellcolor[HTML]{EFEFEF}38.37\%       & \cellcolor[HTML]{EFEFEF}41.83\%   & \cellcolor[HTML]{EFEFEF}0.17   & \cellcolor[HTML]{EFEFEF}5.03    & \cellcolor[HTML]{EFEFEF}706.32 \\
\specialrule{0em}{1pt}{1pt}
(b) & UPAM w/o SPL   & 0.22\%  & 0.37\%  & 0.53         & 5.10          & 707.02 \\ 
(c) & UPAM w/o SEL   & 23.25\% & 28.36\% & 0.17         & 5.06          & 706.58 \\
\bottomrule[0.15em] 
\vspace{-0mm}
\end{tabular}
}
\end{table}
% -----------------------Tab.2
% -----------------------Tab.3
\begin{table}[t]
\centering
\caption{Ablation study on Moving Closer to Boundary (MCB) in SPL.}
\vspace{1mm}
\label{tab:MCB}
\resizebox{0.48\textwidth}{!}{
\begin{tabular}{lcccccc}
\toprule[0.15em] 
&\textbf{Methods}      & \textbf{R-1} ↑    & \textbf{R-3} ↑    & \textbf{Text. Sim.} ↓ & \textbf{Infer. Time} ↓ & \textbf{PPL} ↓  \\ 
\hline
\specialrule{0em}{1pt}{1pt}
\cellcolor[HTML]{EFEFEF}(i) & \cellcolor[HTML]{EFEFEF}UPAM   & \cellcolor[HTML]{EFEFEF}38.37\%       & \cellcolor[HTML]{EFEFEF}41.83\%   & \cellcolor[HTML]{EFEFEF}0.17   & \cellcolor[HTML]{EFEFEF}5.03    & \cellcolor[HTML]{EFEFEF}706.32 \\
\specialrule{0em}{1pt}{1pt}
(ii) & UPAM w/o MCB   & 35.76\% & 39.31\% & 0.22         & 5.03          & 706.10 \\
\bottomrule[0.15em] 
\end{tabular}
}
\end{table}
% -----------------------Tab.3

% -----------------------Tab.4
\begin{table}[t]
\centering
\caption{Ablation study on Gradient Harmonization (GH) in SEL.}
\vspace{1mm}
\label{tab:GH}
\resizebox{0.48\textwidth}{!}{
\begin{tabular}{lcccccc}
\toprule[0.15em] 
&\textbf{Methods}      & \textbf{R-1} ↑    & \textbf{R-3} ↑    & \textbf{Text. Sim.} ↓ & \textbf{Infer. Time} ↓ & \textbf{PPL} ↓  \\ 
\hline
\specialrule{0em}{1pt}{1pt}
\cellcolor[HTML]{EFEFEF}(i) & \cellcolor[HTML]{EFEFEF}UPAM   & \cellcolor[HTML]{EFEFEF}38.37\%       & \cellcolor[HTML]{EFEFEF}41.83\%   & \cellcolor[HTML]{EFEFEF}0.17   & \cellcolor[HTML]{EFEFEF}5.03    & \cellcolor[HTML]{EFEFEF}706.32 \\
\specialrule{0em}{1pt}{1pt}
(ii) & UPAM w/o GH    & 33.82\% & 35.08\% & 0.35         & 5.09          & 706.27 \\
\bottomrule[0.15em] 
\end{tabular}
}
\end{table}
% -----------------------Tab.4

\textbf{Qualitative Results.}  
Qualitative results of our UPAM are shown in Fig. \ref{fig:qualitative}. 
We can see that our generated images are semantically consistent with the ground-truth target images. 
Our adversarial prompts are significantly different from the naive prompts and do not contain any ``harmful'' (malicious) words. Moreover, we can hardly identify the misspelled (unnatural) words, indicating strong attack stealthiness. Qualitative comparisons with existing methods are in \textit{appendix}.

\subsection{Ablation Study}\label{sec:ablation}

To demonstrate the effectiveness of our proposed schemes: SPL and SEL, we conduct ablation experiments by removing the specific SPL (or SEL) in our approach. 
Comparing (a) and (b) in Tab. \ref{tab:abl}, it can be seen that when UPAM does not employ SPL, both R-1 precision and R-3 precision drop to almost 0. 
This is because SPL is designed to deceive textual and visual defenses. In the absence of SPL, black-box T2I models hardly return images, resulting in poor R-precision performance. 
Additionally, SPL also influences the Textual Similarity score, as it can reduce the similarity between adversarial prompts and naive ones, aiming to deceive textual filters within the black-box system.
These ablation results demonstrate the effect of SPL.
When comparing (a) and (c) in Tab. \ref{tab:abl}, we can see that our UPAM without SEL leads to a significant R-precision decrease of 15.12\% (R-1) and 13.47\% (R-3), demonstrating the effect of SEL. We can observe that SEL has virtually no impact on Textual Similarity, Inference Time, and PPL. This is because SEL is proposed to enhance the semantic representation of the returned images, thereby only affecting R-precision.
As SPL and SEL do not alter the LLM-based structure of UPAM, they do not impact the Inference Time and PPL.

\subsection{Effect of further moving closer to the boundary in SPL}
In SPL, after optimizing the model parameters into the ``Pass'' region, we gradually reduce the radius, moving the model parameters closer to the boundary, aiming to increase the likelihood of the generation of target ``harmful'' images, as illustrated in Fig. \ref{fig:spl} (b, c, and d).
To investigate the effect of this design, we conduct ablation by removing this further approach. 
For clarity, we name this design as Moving Closer to Boundary (MCB).
By comparing (i) and (ii) in Tab. \ref{tab:MCB}, when removing MCB in SPL, obvious performance degradation is observed in terms of R-1, R-3, and Textual Similarity.
This demonstrates the effectiveness of further moving closer to the boundary in SPL.

\subsection{Effect of gradient harmonization in SEL}
To improve the compatibility of SEL with SPL, we introduce a Gradient Harmonization (GH) method (see Eq. \ref{eq:sel_gradient}) to adaptively adjust SEL's gradients.
With a comparison between (i) and (ii) in Tab. \ref{tab:GH}, we can see that, when removing GH, SEL could disrupt SPL's optimization achievements, thus leading to worse performances in terms of R-1, R-3, and Textual Similarity.
This demonstrates the effectiveness of gradient harmonization in SEL.

\subsection{Effect of LLM+LoRA}
To ensure the attack stealthiness, we propose to utilize a pre-trained LLM and optimize the LoRA adapter to ensure the naturalness of the generated adversarial prompts.
Here, we conduct two ablation studies: \textbf{(1)} instead of using the pr-trained LLM, we directly employ an untrained transformer \cite{touvron2023llama} (with the same structure as the LLM) and then train it from scratch. 
Comparing (i) and (ii) in Tab. \ref{tab:supp_llm}, we can see the transformer performs much worse in terms of PPL, demonstrating the effect of adopting knowledge of LLM.
\textbf{(2)} Instead of optimizing LoRA, we directly optimize the pre-trained LLM. Comparing (i) and (iii) in Tab. \ref{tab:supp_llm}, the significant change in PPL demonstrates the effect of using LoRA adapter.

% -----------------------Tab.5
\begin{table}[t]
% \vspace{-7mm}
\centering
\caption{Ablation study on LLM and LoRA.}
\vspace{1mm}
\label{tab:supp_llm}
\resizebox{0.48\textwidth}{!}{
\begin{tabular}{lcccccc}
\toprule[0.15em] 
&\textbf{Methods}      & \textbf{R-1} ↑    & \textbf{R-3} ↑    & \textbf{Text. Sim.} ↓ & \textbf{Infer. Time} ↓ & \textbf{PPL} ↓  \\ 
\hline
\specialrule{0em}{1pt}{1pt}
\cellcolor[HTML]{EFEFEF}(i) & \cellcolor[HTML]{EFEFEF}UPAM   & \cellcolor[HTML]{EFEFEF}38.37\%       & \cellcolor[HTML]{EFEFEF}41.83\%   & \cellcolor[HTML]{EFEFEF}0.17   & \cellcolor[HTML]{EFEFEF}5.03    & \cellcolor[HTML]{EFEFEF}706.32 \\
\specialrule{0em}{1pt}{1pt}
(ii) & UPAM w/o LLM   & 38.15\% & 41.68\% & 0.18         & 5.13          & 3954.85 \\
(iii) & UPAM w/o LoRA  & 38.22\% & 41.59\% & 0.17         & 5.05          & 2848.52 \\
\bottomrule[0.15em] 
\end{tabular}
}
\end{table}
% -----------------------Tab.5

\section{Discussion}
In order to prevent malicious use of T2I models to produce harmful or inappropriate content, API providers deploy text filters and visual checkers to prevent the output of harmful images \cite{Midjourney,Leonardo,rombach2022high}. However, as demonstrated in this paper, it is possible to use black-box knowledge to generate imperceptible adversarial prompts for obtaining target images. Therefore, this work further shows the importance of being aware of the security risks of T2I models.

One possible defense mechanism is to finetune the T2I models through \textit{adversarial training} \cite{shafahi2019adversarial}, a common defense method in AI security \cite{goodfellow2014explaining,jin2020bert}, aiming to make the T2I models unable to produce the target image when feeding adversarial prompts. 
However, this approach tends to exhibit a bias towards adversarial prompts used during training, resulting in sub-optimal performance when encountering adversarial prompts from untrained domains \cite{wang2023continual}. 
Another possible defense mechanism is to leverage \textit{model unlearning} \cite{liu2022backdoor} to make T2I models essentially forget harmful knowledge. 
However, it typically requires a significant effort to collect ``unsafe'' images as comprehensively as possible. Moreover, sometimes the scope of harmful images can be difficult to define, making the data collection more labor-intensive and time-consuming \cite{marchant2022hard}.

\section{Conclusion}

In this paper, we introduce UPAM, a unified attack framework for T2I models, designed to deceive both textual filters and visual checkers.
Unlike existing enumeration methods, UPAM enables gradient-based optimization for higher effectiveness and efficiency. 
We devise an SPL scheme to support gradient optimization in the challenging no-feedback scenario, and introduce an SEL scheme to align generated images with desired target semantics.
We incorporate LLM to guarantee attack stealthiness.

\vspace{3mm}
\textbf{\large{Acknowledgements}}

This research/project is supported by the National Research Foundation, Singapore under its AI Singapore Programme (AISG Award No: AISG2-PhD-2022-01-027[T]) and the Ministry of Education, Singapore, under the AcRF Tier 2 Projects (MOE-T2EP20222-0009 and MOE-T2EP20123-0014).

\vspace{3mm}
\textbf{\large{Impact Statement}}

This paper presents work whose goal is to advance the field of Machine Learning. There are many potential social consequences of our work, none of which we feel must be specifically highlighted here. It is important to note that all  experiments are conducted by utilizing natural safe images, with the utmost care to avoid any dissemination of negative consequences. Our research investigates the robustness of text-to-image generation models and highlights the vulnerabilities inherent in these models, and thereby can inform feasible defense approaches for future development in this area. Our study shall be able to encourage the machine learning community to develop trust-worthy, responsible, and robust models. 

\normalem
\bibliography{whole_paper}

\begin{thebibliography}{52}
\providecommand{\natexlab}[1]{#1}
\providecommand{\url}[1]{\texttt{#1}}
\expandafter\ifx\csname urlstyle\endcsname\relax
  \providecommand{\doi}[1]{doi: #1}\else
  \providecommand{\doi}{doi: \begingroup \urlstyle{rm}\Url}\fi

\bibitem[Leo(2023)]{Leonardo}
Leonardo.ai, access date: 9st nov.
\newblock 2023.

\bibitem[Mid(2023)]{Midjourney}
Midjourney, access date: 26th sept.
\newblock 2023.

\bibitem[Birhane et~al.(2021)Birhane, Prabhu, and Kahembwe]{birhane2021multimodal}
Birhane, A., Prabhu, V.~U., and Kahembwe, E.
\newblock Multimodal datasets: misogyny, pornography, and malignant stereotypes.
\newblock \emph{arXiv preprint arXiv:2110.01963}, 2021.

\bibitem[Cer et~al.(2018)Cer, Yang, Kong, Hua, Limtiaco, John, Constant, Guajardo-Cespedes, Yuan, Tar, et~al.]{cer2018universal}
Cer, D., Yang, Y., Kong, S.-y., Hua, N., Limtiaco, N., John, R.~S., Constant, N., Guajardo-Cespedes, M., Yuan, S., Tar, C., et~al.
\newblock Universal sentence encoder.
\newblock \emph{arXiv preprint arXiv:1803.11175}, 2018.

\bibitem[Chesney \& Citron(2019)Chesney and Citron]{chesney2019deep}
Chesney, B. and Citron, D.
\newblock Deep fakes: A looming challenge for privacy, democracy, and national security.
\newblock \emph{Calif. L. Rev.}, 107:\penalty0 1753, 2019.

\bibitem[Chien(2018)]{chien2018source}
Chien, J.-T.
\newblock \emph{Source separation and machine learning}.
\newblock Academic Press, 2018.

\bibitem[Choquette-Choo et~al.(2021)Choquette-Choo, Tramer, Carlini, and Papernot]{choquette2021label}
Choquette-Choo, C.~A., Tramer, F., Carlini, N., and Papernot, N.
\newblock Label-only membership inference attacks.
\newblock In \emph{International conference on machine learning}, pp.\  1964--1974. PMLR, 2021.

\bibitem[Daly et~al.(2020)Daly, Hagendorff, Li, Mann, Marda, Wagner, and Wang]{daly2020ai}
Daly, A., Hagendorff, T., Li, H., Mann, M., Marda, V., Wagner, B., and Wang, W.~W.
\newblock Ai, governance and ethics: global perspectives.
\newblock \emph{University of Hong Kong Faculty of Law Research Paper}, \penalty0 (2020/051), 2020.

\bibitem[Daoud et~al.(2023)Daoud, Shehab, Al-Mimi, Abualigah, Zitar, and Shambour]{daoud2023gradient}
Daoud, M.~S., Shehab, M., Al-Mimi, H.~M., Abualigah, L., Zitar, R.~A., and Shambour, M. K.~Y.
\newblock Gradient-based optimizer (gbo): a review, theory, variants, and applications.
\newblock \emph{Archives of Computational Methods in Engineering}, 30\penalty0 (4):\penalty0 2431--2449, 2023.

\bibitem[Daras \& Dimakis(2022)Daras and Dimakis]{daras2022discovering}
Daras, G. and Dimakis, A.~G.
\newblock Discovering the hidden vocabulary of dalle-2.
\newblock \emph{arXiv preprint arXiv:2206.00169}, 2022.

\bibitem[Finlay et~al.(2019)Finlay, Pooladian, and Oberman]{finlay2019logbarrier}
Finlay, C., Pooladian, A.-A., and Oberman, A.
\newblock The logbarrier adversarial attack: making effective use of decision boundary information.
\newblock In \emph{Proceedings of the IEEE/CVF international conference on computer vision}, pp.\  4862--4870, 2019.

\bibitem[Galetin et~al.(2022)Galetin, {\v{S}}kori{\'c}, and Mihajlovi{\'c}]{galetin2022review}
Galetin, M., {\v{S}}kori{\'c}, J., and Mihajlovi{\'c}, M.
\newblock Review of the standpoints regarding the content of the article 5 of the proposal for the european union's artificial intelligence act: The challenge of finding the balance.
\newblock \emph{Univerzitetska misao-{\v{c}}asopis za nauku, kulturu i umjetnost, Novi Pazar}, \penalty0 (21):\penalty0 164--175, 2022.

\bibitem[Golovin et~al.(2017)Golovin, Solnik, Moitra, Kochanski, Karro, and Sculley]{golovin2017google}
Golovin, D., Solnik, B., Moitra, S., Kochanski, G., Karro, J., and Sculley, D.
\newblock Google vizier: A service for black-box optimization.
\newblock In \emph{Proceedings of the 23rd ACM SIGKDD international conference on knowledge discovery and data mining}, pp.\  1487--1495, 2017.

\bibitem[Goodfellow et~al.(2014)Goodfellow, Shlens, and Szegedy]{goodfellow2014explaining}
Goodfellow, I.~J., Shlens, J., and Szegedy, C.
\newblock Explaining and harnessing adversarial examples.
\newblock \emph{arXiv preprint arXiv:1412.6572}, 2014.

\bibitem[Hu et~al.(2021)Hu, Shen, Wallis, Allen-Zhu, Li, Wang, Wang, and Chen]{hu2021lora}
Hu, E.~J., Shen, Y., Wallis, P., Allen-Zhu, Z., Li, Y., Wang, S., Wang, L., and Chen, W.
\newblock Lora: Low-rank adaptation of large language models.
\newblock \emph{arXiv preprint arXiv:2106.09685}, 2021.

\bibitem[Huang et~al.(2023)Huang, Wu, Liang, Wang, Shi, Wu, Yang, and Zhao]{huang2023towards}
Huang, H., Wu, S., Liang, X., Wang, B., Shi, Y., Wu, P., Yang, M., and Zhao, T.
\newblock Towards making the most of llm for translation quality estimation.
\newblock In \emph{CCF International Conference on Natural Language Processing and Chinese Computing}, pp.\  375--386. Springer, 2023.

\bibitem[Jin et~al.(2020)Jin, Jin, Zhou, and Szolovits]{jin2020bert}
Jin, D., Jin, Z., Zhou, J.~T., and Szolovits, P.
\newblock Is bert really robust? a strong baseline for natural language attack on text classification and entailment.
\newblock In \emph{Proceedings of the AAAI conference on artificial intelligence}, volume~34, pp.\  8018--8025, 2020.

\bibitem[Kahla et~al.(2022)Kahla, Chen, Just, and Jia]{kahla2022label}
Kahla, M., Chen, S., Just, H.~A., and Jia, R.
\newblock Label-only model inversion attacks via boundary repulsion.
\newblock In \emph{Proceedings of the IEEE/CVF conference on computer vision and pattern recognition}, pp.\  15045--15053, 2022.

\bibitem[Kalyan(2024)]{KALYAN2024100048}
Kalyan, K.~S.
\newblock A survey of gpt-3 family large language models including chatgpt and gpt-4.
\newblock \emph{Natural Language Processing Journal}, 6:\penalty0 100048, 2024.

\bibitem[Kim et~al.(2021)Kim, Son, and Kim]{kim2021vilt}
Kim, W., Son, B., and Kim, I.
\newblock Vilt: Vision-and-language transformer without convolution or region supervision.
\newblock In \emph{International Conference on Machine Learning}, pp.\  5583--5594. PMLR, 2021.

\bibitem[Kingma \& Ba(2014)Kingma and Ba]{kingma2014adam}
Kingma, D.~P. and Ba, J.
\newblock Adam: A method for stochastic optimization.
\newblock \emph{arXiv preprint arXiv:1412.6980}, 2014.

\bibitem[Lin et~al.(2014)Lin, Maire, Belongie, Hays, Perona, Ramanan, Doll{\'a}r, and Zitnick]{lin2014microsoft}
Lin, T.-Y., Maire, M., Belongie, S., Hays, J., Perona, P., Ramanan, D., Doll{\'a}r, P., and Zitnick, C.~L.
\newblock Microsoft coco: Common objects in context.
\newblock In \emph{Computer Vision--ECCV 2014: 13th European Conference, Zurich, Switzerland, September 6-12, 2014, Proceedings, Part V 13}, pp.\  740--755. Springer, 2014.

\bibitem[Liu et~al.(2023)Liu, Wu, Zhai, Yuan, and Zhang]{liu2023riatig}
Liu, H., Wu, Y., Zhai, S., Yuan, B., and Zhang, N.
\newblock Riatig: Reliable and imperceptible adversarial text-to-image generation with natural prompts.
\newblock In \emph{Proceedings of the IEEE/CVF Conference on Computer Vision and Pattern Recognition}, pp.\  20585--20594, 2023.

\bibitem[Liu et~al.(2022)Liu, Fan, Chen, Liu, Ma, Wang, and Ma]{liu2022backdoor}
Liu, Y., Fan, M., Chen, C., Liu, X., Ma, Z., Wang, L., and Ma, J.
\newblock Backdoor defense with machine unlearning.
\newblock In \emph{IEEE INFOCOM 2022-IEEE Conference on Computer Communications}, pp.\  280--289. IEEE, 2022.

\bibitem[Marchant et~al.(2022)Marchant, Rubinstein, and Alfeld]{marchant2022hard}
Marchant, N.~G., Rubinstein, B.~I., and Alfeld, S.
\newblock Hard to forget: Poisoning attacks on certified machine unlearning.
\newblock In \emph{Proceedings of the AAAI Conference on Artificial Intelligence}, volume~36, pp.\  7691--7700, 2022.

\bibitem[Milli{\`e}re(2022)]{milliere2022adversarial}
Milli{\`e}re, R.
\newblock Adversarial attacks on image generation with made-up words.
\newblock \emph{arXiv preprint arXiv:2208.04135}, 2022.

\bibitem[Oh et~al.(2023)Oh, Hwang, Lee, Lim, Jung, Jung, Choi, and Song]{oh2023blackvip}
Oh, C., Hwang, H., Lee, H.-y., Lim, Y., Jung, G., Jung, J., Choi, H., and Song, K.
\newblock Blackvip: Black-box visual prompting for robust transfer learning.
\newblock In \emph{Proceedings of the IEEE/CVF Conference on Computer Vision and Pattern Recognition}, pp.\  24224--24235, 2023.

\bibitem[Radford et~al.(2019)Radford, Wu, Child, Luan, Amodei, Sutskever, et~al.]{radford2019language}
Radford, A., Wu, J., Child, R., Luan, D., Amodei, D., Sutskever, I., et~al.
\newblock Language models are unsupervised multitask learners.
\newblock \emph{OpenAI blog}, 1\penalty0 (8):\penalty0 9, 2019.

\bibitem[Radford et~al.(2021)Radford, Kim, Hallacy, Ramesh, Goh, Agarwal, Sastry, Askell, Mishkin, Clark, et~al.]{radford2021learning}
Radford, A., Kim, J.~W., Hallacy, C., Ramesh, A., Goh, G., Agarwal, S., Sastry, G., Askell, A., Mishkin, P., Clark, J., et~al.
\newblock Learning transferable visual models from natural language supervision.
\newblock In \emph{International conference on machine learning}, pp.\  8748--8763. PMLR, 2021.

\bibitem[Raiaan et~al.(2023)Raiaan, Mukta, Fatema, Fahad, Sakib, Mim, Ahmad, Ali, and Azam]{raiaan2023review}
Raiaan, M. A.~K., Mukta, M. S.~H., Fatema, K., Fahad, N.~M., Sakib, S., Mim, M. M.~J., Ahmad, J., Ali, M.~E., and Azam, S.
\newblock A review on large language models: Architectures, applications, taxonomies, open issues and challenges.
\newblock \emph{Authorea Preprints}, 2023.

\bibitem[Ramesh et~al.(2022)Ramesh, Dhariwal, Nichol, Chu, and Chen]{ramesh2022hierarchical}
Ramesh, A., Dhariwal, P., Nichol, A., Chu, C., and Chen, M.
\newblock Hierarchical text-conditional image generation with clip latents.
\newblock \emph{arXiv preprint arXiv:2204.06125}, 1\penalty0 (2):\penalty0 3, 2022.

\bibitem[Rando et~al.(2022)Rando, Paleka, Lindner, Heim, and Tram{\`e}r]{rando2022red}
Rando, J., Paleka, D., Lindner, D., Heim, L., and Tram{\`e}r, F.
\newblock Red-teaming the stable diffusion safety filter.
\newblock \emph{arXiv preprint arXiv:2210.04610}, 2022.

\bibitem[Rombach et~al.(2022)Rombach, Blattmann, Lorenz, Esser, and Ommer]{rombach2022high}
Rombach, R., Blattmann, A., Lorenz, D., Esser, P., and Ommer, B.
\newblock High-resolution image synthesis with latent diffusion models.
\newblock In \emph{Proceedings of the IEEE/CVF conference on computer vision and pattern recognition}, pp.\  10684--10695, 2022.

\bibitem[Saharia et~al.(2022)Saharia, Chan, Saxena, Li, Whang, Denton, Ghasemipour, Gontijo~Lopes, Karagol~Ayan, Salimans, et~al.]{saharia2022photorealistic}
Saharia, C., Chan, W., Saxena, S., Li, L., Whang, J., Denton, E.~L., Ghasemipour, K., Gontijo~Lopes, R., Karagol~Ayan, B., Salimans, T., et~al.
\newblock Photorealistic text-to-image diffusion models with deep language understanding.
\newblock \emph{Advances in Neural Information Processing Systems}, 35:\penalty0 36479--36494, 2022.

\bibitem[Schramowski et~al.(2023)Schramowski, Brack, Deiseroth, and Kersting]{schramowski2023safe}
Schramowski, P., Brack, M., Deiseroth, B., and Kersting, K.
\newblock Safe latent diffusion: Mitigating inappropriate degeneration in diffusion models.
\newblock In \emph{Proceedings of the IEEE/CVF Conference on Computer Vision and Pattern Recognition}, pp.\  22522--22531, 2023.

\bibitem[Shafahi et~al.(2019)Shafahi, Najibi, Ghiasi, Xu, Dickerson, Studer, Davis, Taylor, and Goldstein]{shafahi2019adversarial}
Shafahi, A., Najibi, M., Ghiasi, M.~A., Xu, Z., Dickerson, J., Studer, C., Davis, L.~S., Taylor, G., and Goldstein, T.
\newblock Adversarial training for free!
\newblock \emph{Advances in Neural Information Processing Systems}, 32, 2019.

\bibitem[Spall(1992{\natexlab{a}})]{Sp1992multi}
Spall, J.
\newblock Multivariate stochastic approximation using a simultaneous perturbation gradient approximation.
\newblock \emph{IEEE Transactions on Automatic Control}, 37\penalty0 (3):\penalty0 332--341, 1992{\natexlab{a}}.

\bibitem[Spall(1992{\natexlab{b}})]{spall1992multivariate}
Spall, J.~C.
\newblock Multivariate stochastic approximation using a simultaneous perturbation gradient approximation.
\newblock \emph{IEEE transactions on automatic control}, 37\penalty0 (3):\penalty0 332--341, 1992{\natexlab{b}}.

\bibitem[Spall(1997{\natexlab{a}})]{spall1997accelerated}
Spall, J.~C.
\newblock Accelerated second-order stochastic optimization using only function measurements.
\newblock In \emph{Proceedings of the 36th IEEE Conference on Decision and Control}, volume~2, pp.\  1417--1424. IEEE, 1997{\natexlab{a}}.

\bibitem[Spall(1997{\natexlab{b}})]{spall1997one}
Spall, J.~C.
\newblock A one-measurement form of simultaneous perturbation stochastic approximation.
\newblock \emph{Automatica}, 33\penalty0 (1):\penalty0 109--112, 1997{\natexlab{b}}.

\bibitem[Spall(2000)]{spall2000adaptive}
Spall, J.~C.
\newblock Adaptive stochastic approximation by the simultaneous perturbation method.
\newblock \emph{IEEE transactions on automatic control}, 45\penalty0 (10):\penalty0 1839--1853, 2000.

\bibitem[Spall(2005)]{spall2005introduction}
Spall, J.~C.
\newblock \emph{Introduction to stochastic search and optimization: estimation, simulation, and control}.
\newblock John Wiley \& Sons, 2005.

\bibitem[Stephan et~al.(2017)Stephan, Hoffman, Blei, et~al.]{stephan2017stochastic}
Stephan, M., Hoffman, M.~D., Blei, D.~M., et~al.
\newblock Stochastic gradient descent as approximate bayesian inference.
\newblock \emph{Journal of Machine Learning Research}, 18\penalty0 (134):\penalty0 1--35, 2017.

\bibitem[Struppek et~al.(2022)Struppek, Hintersdorf, and Kersting]{struppek2022biased}
Struppek, L., Hintersdorf, D., and Kersting, K.
\newblock The biased artist: Exploiting cultural biases via homoglyphs in text-guided image generation models.
\newblock \emph{arXiv preprint arXiv:2209.08891}, 2022.

\bibitem[Touvron et~al.(2023)Touvron, Lavril, Izacard, Martinet, Lachaux, Lacroix, Rozi{\`e}re, Goyal, Hambro, Azhar, et~al.]{touvron2023llama}
Touvron, H., Lavril, T., Izacard, G., Martinet, X., Lachaux, M.-A., Lacroix, T., Rozi{\`e}re, B., Goyal, N., Hambro, E., Azhar, F., et~al.
\newblock Llama: Open and efficient foundation language models.
\newblock \emph{arXiv preprint arXiv:2302.13971}, 2023.

\bibitem[Wang et~al.(2023)Wang, Liu, Ling, Li, Liu, Li, Chen, Yuille, and Yu]{wang2023continual}
Wang, Q., Liu, Y., Ling, H., Li, Y., Liu, Q., Li, P., Chen, J., Yuille, A., and Yu, N.
\newblock Continual adversarial defense.
\newblock \emph{arXiv preprint arXiv:2312.09481}, 2023.

\bibitem[Xu et~al.(2018)Xu, Zhang, Huang, Zhang, Gan, Huang, and He]{xu2018attngan}
Xu, T., Zhang, P., Huang, Q., Zhang, H., Gan, Z., Huang, X., and He, X.
\newblock Attngan: Fine-grained text to image generation with attentional generative adversarial networks.
\newblock In \emph{Proceedings of the IEEE conference on computer vision and pattern recognition}, pp.\  1316--1324, 2018.

\bibitem[Yu et~al.(2022)Yu, Xu, Koh, Luong, Baid, Wang, Vasudevan, Ku, Yang, Ayan, et~al.]{yu2022scaling}
Yu, J., Xu, Y., Koh, J.~Y., Luong, T., Baid, G., Wang, Z., Vasudevan, V., Ku, A., Yang, Y., Ayan, B.~K., et~al.
\newblock Scaling autoregressive models for content-rich text-to-image generation.
\newblock \emph{arXiv preprint arXiv:2206.10789}, 2\penalty0 (3):\penalty0 5, 2022.

\bibitem[Zhang et~al.(2017)Zhang, Xu, Li, Zhang, Wang, Huang, and Metaxas]{zhang2017stackgan}
Zhang, H., Xu, T., Li, H., Zhang, S., Wang, X., Huang, X., and Metaxas, D.~N.
\newblock Stackgan: Text to photo-realistic image synthesis with stacked generative adversarial networks.
\newblock In \emph{Proceedings of the IEEE international conference on computer vision}, pp.\  5907--5915, 2017.

\bibitem[Zhang et~al.(2018)Zhang, Xu, Li, Zhang, Wang, Huang, and Metaxas]{zhang2018stackgan++}
Zhang, H., Xu, T., Li, H., Zhang, S., Wang, X., Huang, X., and Metaxas, D.~N.
\newblock Stackgan++: Realistic image synthesis with stacked generative adversarial networks.
\newblock \emph{IEEE transactions on pattern analysis and machine intelligence}, 41\penalty0 (8):\penalty0 1947--1962, 2018.

\bibitem[Zhou et~al.(2021)Zhou, Wang, Heidari, Zhao, Turabieh, and Chen]{zhou2021random}
Zhou, W., Wang, P., Heidari, A.~A., Zhao, X., Turabieh, H., and Chen, H.
\newblock Random learning gradient based optimization for efficient design of photovoltaic models.
\newblock \emph{Energy Conversion and Management}, 230:\penalty0 113751, 2021.

\bibitem[Zhu et~al.(2022)Zhu, Ye, Zhou, Liu, and Zhou]{zhu2022label}
Zhu, T., Ye, D., Zhou, S., Liu, B., and Zhou, W.
\newblock Label-only model inversion attacks: Attack with the least information.
\newblock \emph{IEEE Transactions on Information Forensics and Security}, 18:\penalty0 991--1005, 2022.

\end{thebibliography}
\bibliographystyle{icml2024}

\newpage
\appendix
\onecolumn

%%%%%%%%%%%%%%%%%%%%%%%%%%%%%%%%%%%%%%%%%%%%%%%%%%%%%%%%%%%%%%%%%%%%%%%%%%%%%%%
%%%%%%%%%%%%%%%%%%%%%%%%%%%%%%%%%%%%%%%%%%%%%%%%%%%%%%%%%%%%%%%%%%%%%%%%%%%%%%%
% APPENDIX
%%%%%%%%%%%%%%%%%%%%%%%%%%%%%%%%%%%%%%%%%%%%%%%%%%%%%%%%%%%%%%%%%%%%%%%%%%%%%%%
%%%%%%%%%%%%%%%%%%%%%%%%%%%%%%%%%%%%%%%%%%%%%%%%%%%%%%%%%%%%%%%%%%%%%%%%%%%%%%%

\section{Details of Evaluation Metrics}

We evaluate attack methods from the following perspectives:

\textbf{Attack Effectiveness.}
Following previous work \cite{liu2023riatig}, we measure the attack effectiveness by evaluating the generated adversarial prompts $\mathbf{T}^{\ast}$ and the returned images $\mathbf{I}^{\ast}$, respectively.
(1) As for the returned images $\mathbf{I}^{\ast}$, we follow \cite{liu2023riatig} to utilize \uline{R-precision} to measure the attack effectiveness. 
Specifically, a retrieval experiment is conducted using the returned image to query against a set of candidate text descriptions. 
This set comprises 1 ground-truth description and 99 randomly chosen mismatched descriptions. 
For this purpose, a state-of-the-art image-text retrieval model \cite{kim2021vilt} is employed here. 
We collect the top $R$ ranked results in the retrieval, deeming our attacks successful if the ground truth description is among them. We follow \cite{liu2023riatig} to conduct evaluation at R-1 precision (i.e., $R=1$) and R-3 precision (i.e., $R=3$).
Note that if no image is returned to the attacker, the R-precision score of this sample is 0.
(2) As for the generated adversarial prompts $\mathbf{T}^{\ast}$, we use \uline{textual similarity} to evaluate the attack effectiveness.
The effective adversarial prompts should show low textual similarity with naive prompts. 
Therefore, following previous work \cite{liu2023riatig,jin2020bert}, we adopt the Universal Sentence Encoder \cite{cer2018universal} to encode the adversarial prompts $\mathbf{T}^{\ast}$ and naive prompts $\mathbf{T}$ into high dimensional vectors, and then compute their textual similarity score. 
The lower the textual similarity, the better the attack.

\textbf{Attack Efficiency.}
We also evaluate the attack efficiency of the methods. 
Specifically, we present the \uline{average inference time} taken for each testing sample during testing.
Due to the uncertain inference time of enumeration methods, in order to avoid endless waiting periods, we set a criterion: if the inference time exceeds 10 minutes for a given sample, the attack is considered unsuccessful, and the inference process is stopped.

\textbf{Attack Stealthiness.} 
To evaluate the attack stealthiness, we follow \cite{liu2023riatig} to evaluate the naturalness of the generated adversarial prompts by computing the \uline{perplexity score (PPL)} using GPT-2 \cite{radford2019language}, a model trained on extensive real-world sentences. 
Generally, the prompt sample with a lower PPL is more natural.

\section{Modification of UPAM for Protocol B}
\begin{figure}[ht]
\centering
\includegraphics[width=0.9\textwidth]{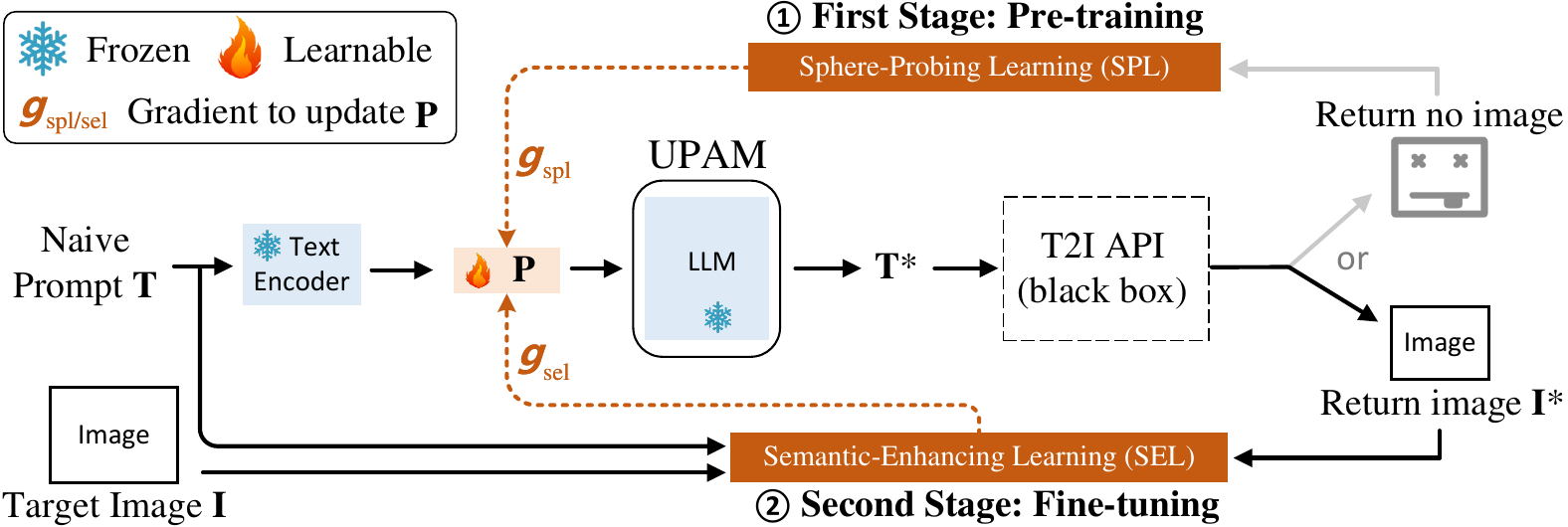} 
\vspace{-0mm}
\caption{Overview of our UPAM framework modified for Protocol B.
}
\vspace{-3mm}
\label{fig:pipeline_for_B}
\end{figure}

As for previous enumeration-based methods, given a naive prompt $\mathbf{T}$ and a target image $\mathbf{I}$, instead of training parameterized models, they aim to directly find an adversarial prompt $\mathbf{T}^{\ast}$ based on the given data $\mathbf{T}$ and $\mathbf{I}$. 
We refer to this as Protocol B. 
To ensure a fair comparison with previous enumeration-based methods, we make slight modifications to UPAM to enable it to directly find adversarial prompts without learning from the training set. 
The modified architecture is illustrated in Fig. \ref{fig:pipeline_for_B}. Next, we provide modification details as follows:

1. As for the structure of our UPAM, we no longer utilize the LoRA adapter while only adopting the LLM. The parameters of the LLM remain frozen.

2. Given $\{ \mathbf{T}, \mathbf{I}  \}$, following the processes of SPL and SEL of our main paper, we still compute gradients $\pmb{g}_\mathrm{spl}$ and $\pmb{g}_\mathrm{sel}$. The difference is that, these gradients no longer optimize LoRA parameters but optimize the input soft embedding $\mathbf{P}$. The soft embedding $\mathbf{P}$ is obtained by feeding the naive $\mathbf{T}$ into the text encoder of LLM. 

In this way, given a naive prompt $\mathbf{T}$ and a target image $\mathbf{I}$, we can directly find a proper adversarial prompt $\mathbf{T}^{\ast}$ by optimizing the input embedding $\mathbf{P}$, without training any model parameters.
\vspace{3mm}

% ==============================Figure 6
% ==============================Figure 6

\section{More Qualitative Results}
Here, we present additional qualitative results. Specifically, we compare our UPAM with  HiddVocab \cite{daras2022discovering}, MacPromp \cite{milliere2022adversarial}, and RIATIG  \cite{liu2023riatig}, while not showcasing the results of TextFooler \cite{jin2020bert}, HomoSubs \cite{struppek2022biased}, and EvoPromp \cite{milliere2022adversarial}. This is because the latter three methods can hardly compel the black-box API to return images, and thus their results cannot be presented.

\begin{figure*}[t]
\centering
\includegraphics[width=0.99\textwidth]{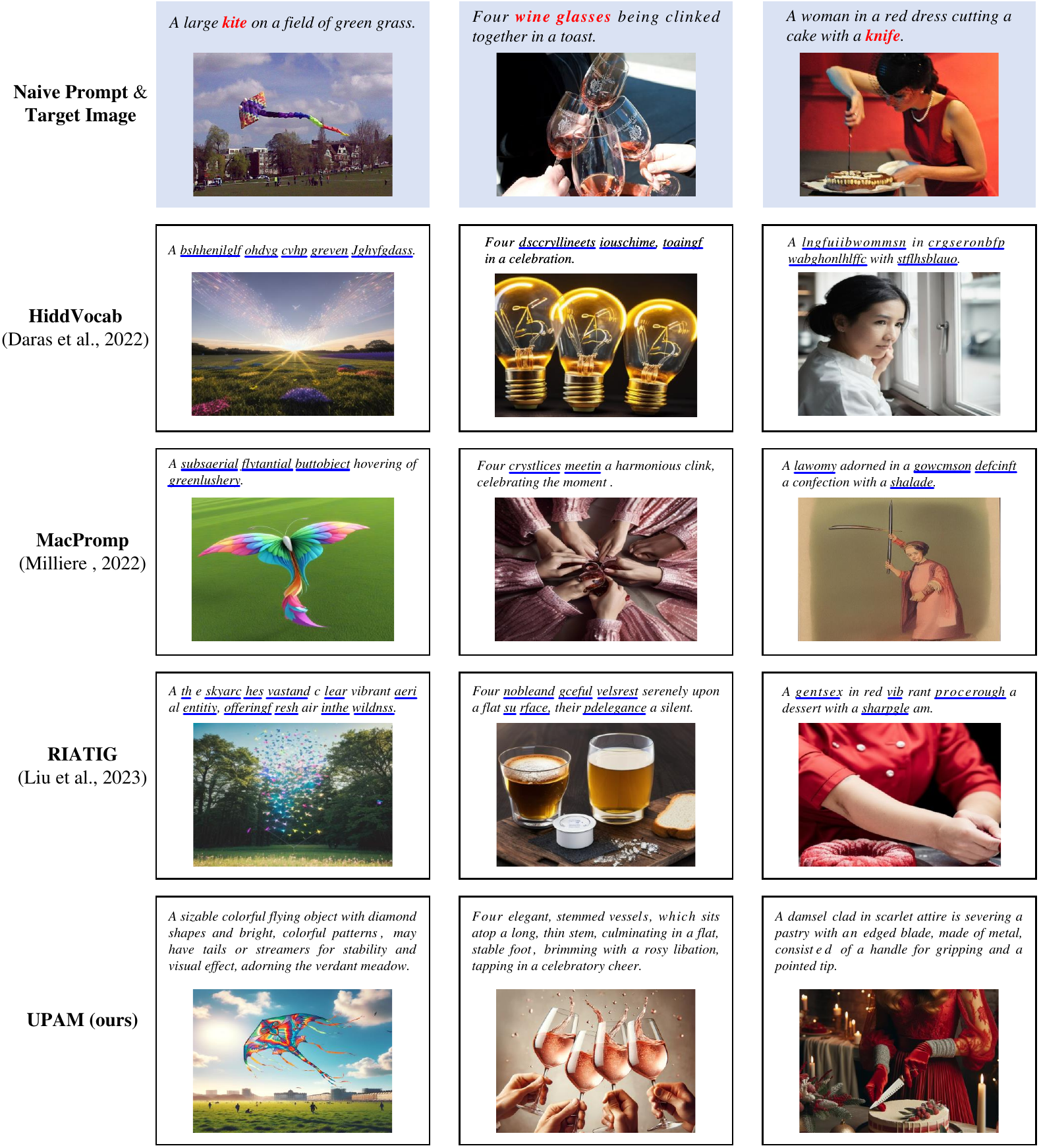} 
\caption{Qualitative comparison results with existing methods. In each block, the text above represents the adversarial prompts generated by the corresponding method, while the pictures below show the images returned from the T2I model.
The name of each ``harmful'' class is marked in \textcolor{red}{red}. We employ \textcolor{blue}{blue underlines} to show unnatural words within adversarial prompts.
}
\label{fig:qua_1}
\end{figure*}

In Fig. \ref{fig:qua_1}, we showcase the attack results. We can see that, in order to bypass textual filters, all methods generate adversarial prompts that no longer contain the sensitive words of naive prompts (marked in red). However, existing methods produce adversarial text with noticeably unnatural words (highlighted with blue underlines). In contrast, our method generates adversarial prompts that contain no misspelled words and maintain grammatical correctness, thereby exhibiting superior naturalness. Moreover, compared to other methods, our UPAM achieves the closest semantic alignment between the generated images and target images.

% ==============================Figure 7
% ==============================Figure 7

\end{document}